\documentclass[10pt,a4paper]{article}
\usepackage[utf8]{inputenc}

\usepackage{xspace}
\usepackage{color}
\usepackage{enumerate}
\usepackage{amsmath}
\usepackage{amssymb}
\usepackage{amsthm}
\usepackage{hyperref}
\usepackage{cleveref}
\usepackage{mathtools}
\usepackage{graphicx}

\usepackage{listings}

\usepackage{xcolor, soul}
\usepackage{subcaption}

\usepackage{float}

\usepackage[margin=1in]{geometry}

\usepackage[tt=false]{libertine}  %
\usepackage[]{sourcecodepro}

\usepackage{todonotes}   %

\Crefname{figure}{Figure}{Figures}
\crefname{figure}{figure}{figures}

\input Sanremo.fd

\newcommand{\mfst}{{\usefont{U}{Sanremo}{xl}{n} M}FST\xspace}

\title{\mfst: A Python OpenFST Wrapper With Support for Custom Semirings and Jupyter Notebooks}
\author{Matthew Francis-Landau \\
  {\small\tt mfl@cs.jhu.edu}}

\date{}

\begin{document}

\maketitle

\begin{abstract}
  \noindent
  This paper introduces \mfst, a new Python library for working with
  Finite-State Machines based on OpenFST.  \mfst is a thin wrapper for OpenFST
  and exposes all of OpenFST's methods for manipulating FSTs.  Additionally,
  \mfst is the only Python wrapper for OpenFST that exposes OpenFST's ability
  to define a \emph{custom} semirings.  This makes \mfst ideal for
  developing models that involve learning the weights on a FST or creating
  \emph{neuralized} FSTs.  \mfst has been designed to be easy to get started with and
  has been previously used in homework assignments for a NLP class as well in
  projects for integrating FSTs and neural networks.  In this paper, we exhibit
  \mfst API and how to use \mfst to build a simple neuralized FST with PyTorch.
\end{abstract}

\section{Finite-State Machine}

\begin{figure}[h]
\centering
\includegraphics[width=.8\textwidth]{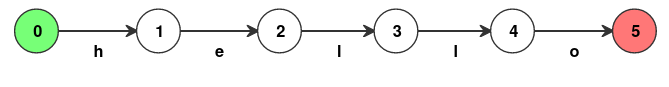}
\caption{Finite-State Acceptor (FSA) for the string \emph{``hello.''}  State \#0, is
  colored green as it is the initial state of the FSA.  State \#5 is
  marked red as it is a terminal state.  Strings that do not exactly match
  \emph{``hello''} will not reach the final state (red), thus are not accepted.}
\label{fig:fsa-hello}
\end{figure}

\begin{figure}[h]
  \centering
  \hspace{-10mm}
  \includegraphics[width=.8\textwidth,trim=120 0 100 450,clip]{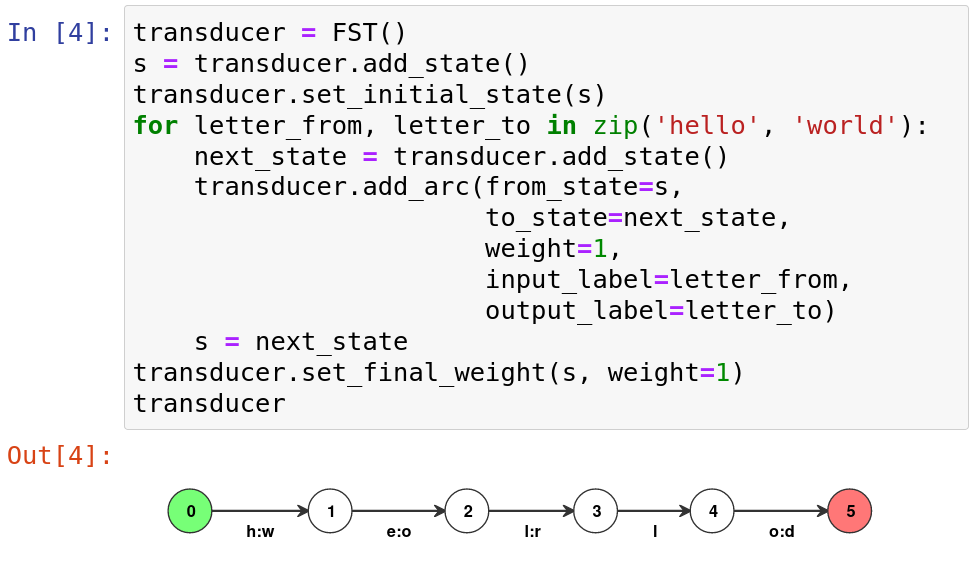}
  \caption{Finite-State Transducer (FST) that transforms the string \emph{``hello''}
    $\to$ \emph{``world.''}  During each transition, a character of the string
    \emph{``hello''} is read from the input string---represented on the left side of the
    colon.  On the right hand side of the colon, the output ``world'' is generated one character at a time as each arc is transversed.  When the
    characters are the same on both the input and output side (e.g. between
    states $3 \to 4$), the colon is omitted for readability.}
  \label{fig:fst-hello-world}
\end{figure}

\begin{figure}[h]
  \centering
  \includegraphics[width=.8\textwidth,trim=120 0 80 555,clip]{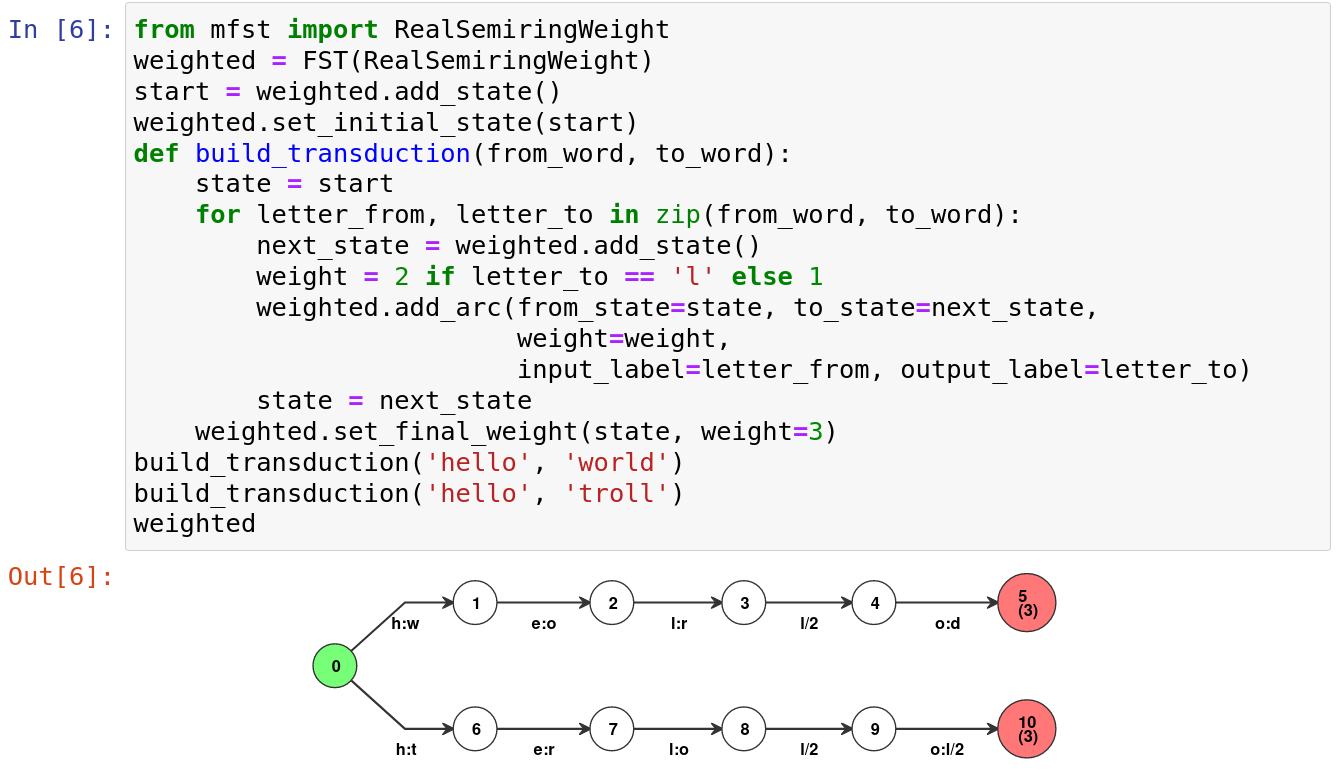}

  \caption{Weighted Finite-State Transducer where \emph{``hello''} is transduced
    non-deterministically into both the words \emph{``world''} and \emph{``troll.''}  Each
    output word is weighted with the product of the weights along a given path.
    Here, \emph{``hello''} is given the weight $1*1*1*2*1*3 = 6$, and \emph{``troll''} is
    given $1*1*1*2*2*3 = 12$.  The weight $1$ is omitted from the figure for
    readability.  If there exist multiple paths that generate the same output,
    the weight would be summed across the different paths.}
  \label{fig:wfst-hello-world-troll}
\end{figure}

A Finite-State Machine (FSM) represents a computation as a finite number of states
labeled $[0, N)$ and transitions between those states.  A FSM starts at an
initial state---represented as {\sethlcolor{green}\hl{green}} in
this paper---and performs a series of transitions following the directed edges
until it reaches a final state---represented as {\sethlcolor{red}\hl{red}} in
this paper.  Transitions in a FSM can be labeled with symbols (e.g. a character
in a string).  In general, a Finite-State Transducer (FST) contains two symbols
on each edge.  One symbol represents what is \emph{read} from the FST's input,
and the other represents what is \emph{written} to the output
(\cref{fig:fst-hello-world}).  In the special case that both symbols are the
same along all edges, a FST can also be called a Finite-State Acceptor (FSA)
(\cref{fig:fsa-hello}).  Furthermore, edges in a FST can also be weighted.
This allows for a FST to score different paths that accept a given string. 
If an input can be mapped to multiple outputs, then each output will have a
different weight assigned (\cref{fig:wfst-hello-world-troll}).

In the remainder of this paper, we will assume the reader is familiar with the theory and
algorithms behind FSTs.  Instead, we will focus on \mfst and how it makes working with
FSTs from Python easy.  The examples in this paper are intended to demonstrate
how easy it is to get started constructing simple models rather than
demonstrating ``real'' models that work with real scenarios.

\section{\mfst: Getting Started}
\mfst was initially developed for teaching advanced FST techniques in a classroom
setting~\cite{seq2class_assignment2}.  This has motivated \mfst's simple interface,
which makes it quick to get started.  \mfst includes sensible defaults all
accessible from Python, while not eliminating power user features of OpenFST.

\subsection{One Command Install}
\mfst is installable with a single command.  \mfst includes and automatically
compiles OpenFST, requiring no additional steps from a user\footnote{Note: The
  install command takes around 10 minutes to run without printing any indication
  it is running to the terminal.}.
\begin{lstlisting}
pip install 'git+https://github.com/matthewfl/openfst-wrapper.git'
\end{lstlisting}

\subsection{First Simple FSAs}
Once \mfst is installed, it makes the first steps of interacting with a FSTs from
Python easy.  Here, a FSA is constructed for the string \emph{``hello.''}  The FSA is
constructed using \mfst's default semiring---more on that later in \cref{sec:weighted-fsts}.

\begin{figure}[h]
  \centering
  \includegraphics[width=.8\textwidth]{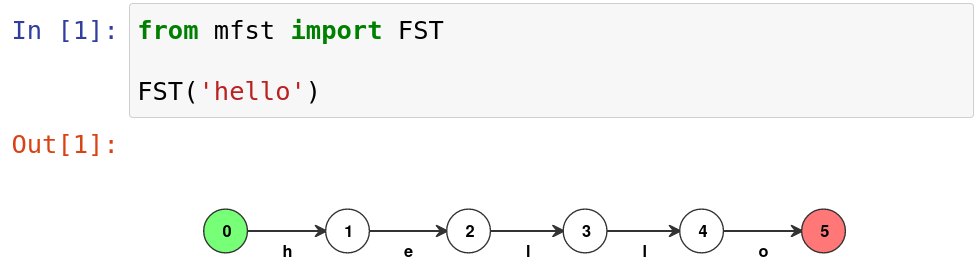}
  \caption{The simplest way to construct a FSA for the string \emph{``hello.''}  When
    a FST is returned from a Jupyter notebook cell, it is drawn into the
    notebook.  All figures in this paper are from \mfst's Jupyter drawings.
    Here we are exemplifying how any Python iterable (such as a string) is
    converted into the labels on a FSA.}
\label{fig:fsa-hello-code}
\end{figure}

\begin{figure}[h]
  \centering
  \includegraphics[width=.8\textwidth]{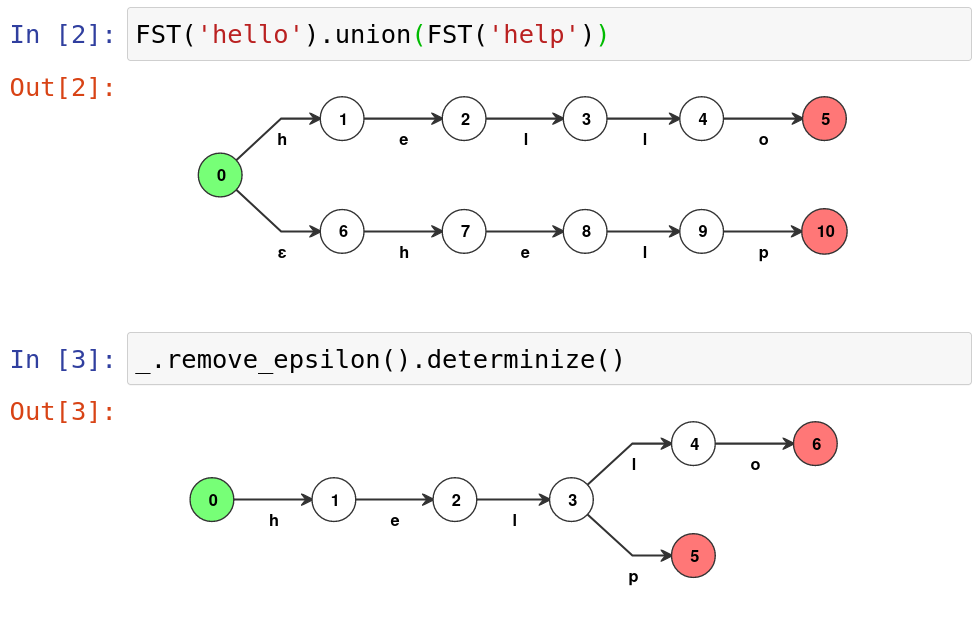}
  \caption{Simple FSA constructed by taking the union of the FSA for the string
    \emph{``hello''} and \emph{``help.''}  In cell \#3, the FSA is referenced using Pythons
    underscore variable (\texttt{\_}) to reference the output of the previous
    computation.  The operations \texttt{remove\_epsilon} and
    \texttt{determinize} are chained together here to create a FSA where the
    prefixes \emph{``hel''} is merged.}
  \label{fig:union-determinize}
\end{figure}

\begin{figure}[h]
  \centering
  \begin{subfigure}[b]{.62\textwidth}
    \includegraphics[width=\textwidth,trim=0 200 400 0,clip]{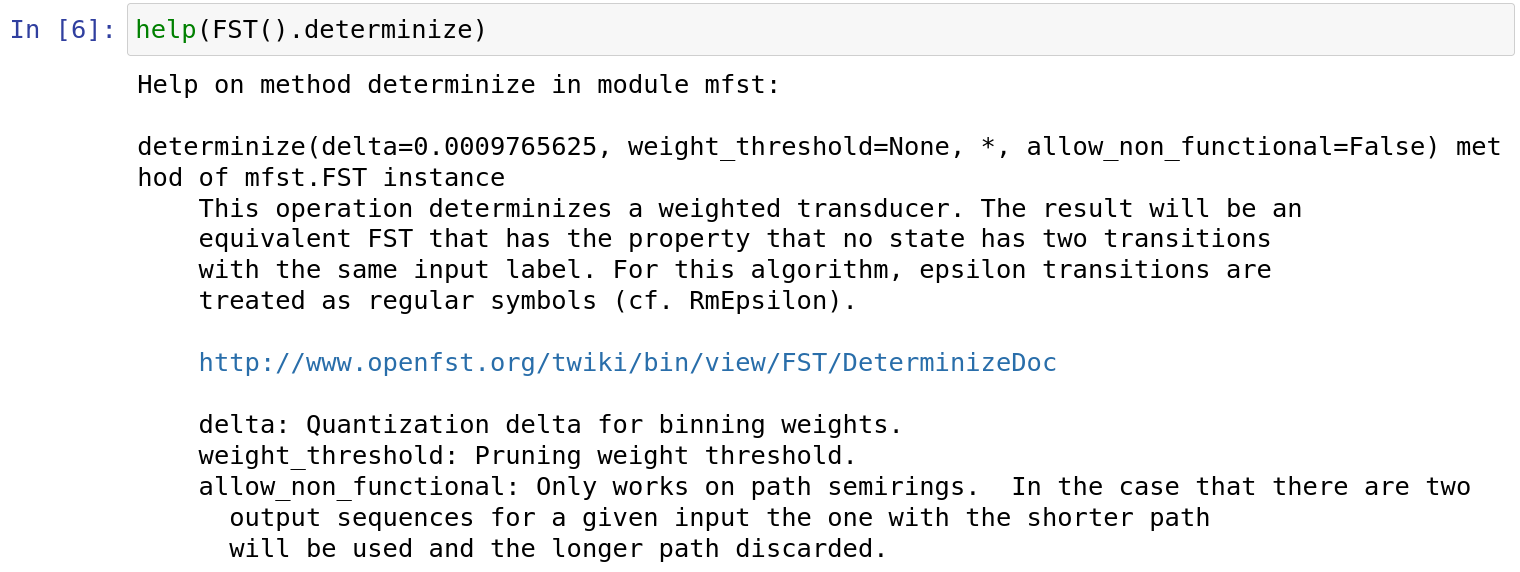}
    \caption{Using Python's \texttt{help()} function to access documentation for a
      method.  The documentation is largely adapted from OpenFST online
      documentation$^{\text{\ref{footnote:openfst_docs}}}$.}
  \end{subfigure}%
  \hfill
  \begin{subfigure}[b]{.35\textwidth}
    \includegraphics[width=\textwidth,trim=0 200 150 0,clip]{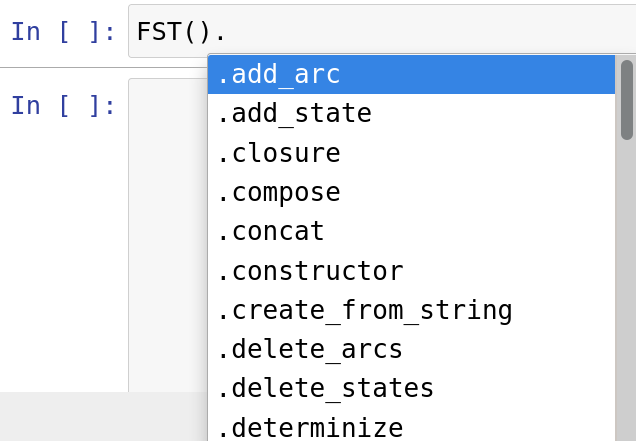}
    \caption{Tab competition in a Jupyter notebook to list the available methods
      on the FST class.}
  \end{subfigure}
  \caption{The documentation on \mfst methods is included as Python docstrings
    and can also be easily auto-completed using any
    Python IDE.}
  \label{fig:docs-included}
\end{figure}

\mfst exposes all of OpenFST's
operations\footnote{OpenFST included operations:
  \url{http://www.openfst.org/twiki/bin/view/FST/FstQuickTour\#Available\%20FST\%20Operations}.\label{footnote:openfst_docs}}
for interacting and manipulating a FST.  Whenever an instance of the
\texttt{FST} class is returned from a cell in a Jyputer notebook, the FST is
automatically drawn into the notebook.  This makes \mfst ideal for learning how
FSTs work, visualizing how the different operations transform a FST without
requiring any additional steps to draw the FST (shown in
\cref{fig:union-determinize}).  \mfst, being a Python package, includes
documentation on all of its methods (adapted from OpenFST's documentation).
This documentation can be easily accessed in an interactive Python environment
using Python's \texttt{help()} method (\cref{fig:docs-included}).

\section{Building Transducers}
\begin{figure}[h]
  \centering
  \includegraphics[width=.8\textwidth]{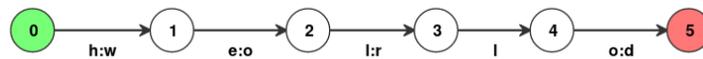}

  \caption{In Jupyter cell \#4, an empty FST is first constructed using
    \texttt{FST()} with the default semiring.  The edges are added one at a time
    using \texttt{add\_arc}, where most edges have different labels on the input
    and output side.  }
  \label{fig:build-fst}
\end{figure}

FST can also be built using the \texttt{add\_state} and \texttt{add\_arc}
methods in addition to the simplified constructor we have demonstated so far.  These methods
makes it possible to build more complicated FSTs which may have loops or may define
input and output labels and weights on all of the edges.  The main methods for
constructing a FST are as follows, and are shown in \cref{fig:build-fst}.

\begin{itemize}
\item \texttt{add\_state()$\to$state\_id}: Modifies the FST in place to add a new state.

\item \texttt{add\_arc(from\_state, to\_state, weight, input\_label, output\_label)$\to$unit}:\linebreak Modifies the FST in place to add a new arc between two states.  If \texttt{from\_state} and \texttt{to\_state} are the same, then this will create a self-loop.  The \texttt{weight} is cast to the FST's semiring automatically.  By default \texttt{weight} is set to the semiring's identity element of $\bar{1}$.  \texttt{input\_label} and \texttt{output\_label} are 64 bit integers.  The value $0$ is used to represent the special $\epsilon$ character in the FST, which indicates that no symbol is read/written when traversing an arc.  If a single character of a string is passed, then it will be automatically converted into an integer using it character code (the method \texttt{chr} in Python).

\item \texttt{set\_initial\_state(state\_id)$\to$unit}: Modifies the FST to set its initial state.  A FST can only have a single initial state, and a FST without an initial state will not be usable.

\item \texttt{set\_final\_weight(state\_id, weight)$\to$unit}: Modifies the FST to set the final state weight.  There can be multiple final states, and a well defined FST should have at least one final state.
\end{itemize}

\begin{figure}[h]
\centering
\includegraphics[width=.8\textwidth]{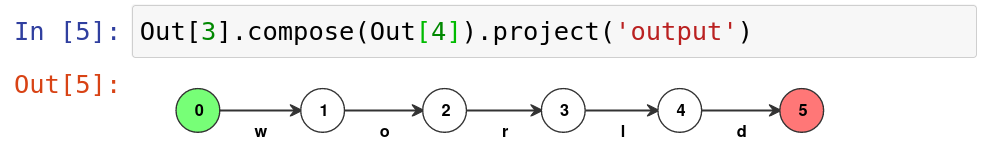}
\caption{In cell \#5, the FST is composed with the FSA from
    \cref{fig:union-determinize} that was the union of the word \emph{``hello''} and
    \emph{``help.''}  Only the word \emph{``hello''} is transduced as there is no path that
    accepts the word \emph{``help.''}  The \texttt{project('output')} method turns the
    FST into a FSA using the labels on the output side.}
\end{figure}

\section{Weighted FSTs} \label{sec:weighted-fsts}

\begin{figure}[h]
  \centering
  \begin{subfigure}[t]{.48\textwidth}
    \includegraphics[width=\textwidth]{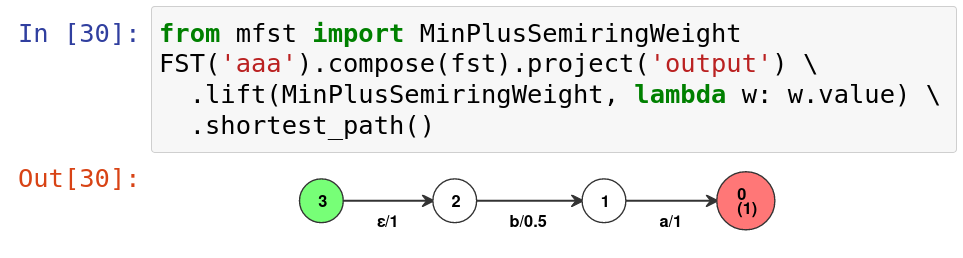}
  \end{subfigure}
  \begin{subfigure}[t]{.48\textwidth}
    \includegraphics[width=\textwidth]{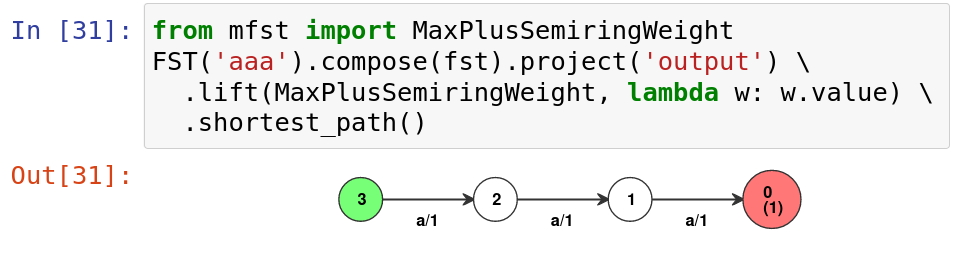}
  \end{subfigure}
  \caption{The semiring is cast to the min-plus and max-plus semirings and then
    we use the shortest path algorithm to identify the shortest path.  In these
    semirings, the multiplication ($\odot$) operator is summing the length of the path
    and the addition ($\oplus$) operator is defined as min/max.  This allows us to find the shortest
    path using the ``$\le$'' operator where it is defined as
    $a \le b \iff a \oplus b = a$ for idempotent semirings.}
  \label{fig:cast-fst}
\end{figure}

Weighted FSTs allow for scoring a given sequence rather than just accepting or
transducing an input.  This is done by assigning a weight to every edge in the
FST.  The weights on a FST must all be instances of the same semiring, meaning
that they share a multiplication and addition operator as well as a multiplicative and
additive identity $\langle +, *, \bar{0}, \bar{1} \rangle$.
To determine the weight for a given input/output sequence, the weights along a
given path are multiplied together.  When there are multiple paths that accept
an input or transduce a given input/output pair, the weight will be summed
across all paths.  Changing the semiring and defining different actions for
multiplication and addition can allow the same algorithm operating on the FST to
compute different results---such as in \cref{fig:cast-fst}.

\mfst includes the standard semirings.  This includes:
\href{https://github.com/matthewfl/openfst-wrapper/blob/dea1c93741b0c3be3dcedaa5d6f9158ce4126453/mfst/semirings.py#L15}{Python values},
\href{https://github.com/matthewfl/openfst-wrapper/blob/dea1c93741b0c3be3dcedaa5d6f9158ce4126453/mfst/semirings.py#L85}{real numbers},
\href{https://github.com/matthewfl/openfst-wrapper/blob/dea1c93741b0c3be3dcedaa5d6f9158ce4126453/mfst/semirings.py#L112}{min},
\href{https://github.com/matthewfl/openfst-wrapper/blob/dea1c93741b0c3be3dcedaa5d6f9158ce4126453/mfst/semirings.py#L140}{max},
 \href{https://github.com/matthewfl/openfst-wrapper/blob/dea1c93741b0c3be3dcedaa5d6f9158ce4126453/mfst/semirings.py#L168}{tropical}, and
\href{https://github.com/matthewfl/openfst-wrapper/blob/dea1c93741b0c3be3dcedaa5d6f9158ce4126453/mfst/semirings.py#L178}{boolean}.
Each of these semirings is implemented in approximately 20 lines of Python each.
Custom semirings can be defined by extending the
\href{https://github.com/matthewfl/openfst-wrapper/blob/dea1c93741b0c3be3dcedaa5d6f9158ce4126453/mfst/__init__.py#L16}{\texttt{AbstractSemiringWeight}}
class.

\mfst's default semiring---that we have been using so far---is the boolean
semiring.  In the boolean semiring, all weights along the edges are $\bar{1}$
(as the $\bar{0}$ weights are omitted).  The boolean semiring is given
special precedence in \mfst as the boolean semiring will be automatically cast
to another semiring as necessary.  All operations in \mfst which involve two or
more FSTs---such as compose or union---require that all of its arguments
implement the same semiring.  The boolean semiring's automatic casting is useful
as it allows us to construct FSTs and compose them with weighted FSTs without
too much concern for which semiring is in use, such as in
\cref{fig:compose-aaa}.  \mfst also provides the method
\texttt{lift(semiring\_class, cast\_method)$\to$FST} to convert a FST from one
semiring to another, such as in \cref{fig:cast-fst}.

\begin{figure}[H]
  \centering
  \includegraphics[width=.8\textwidth]{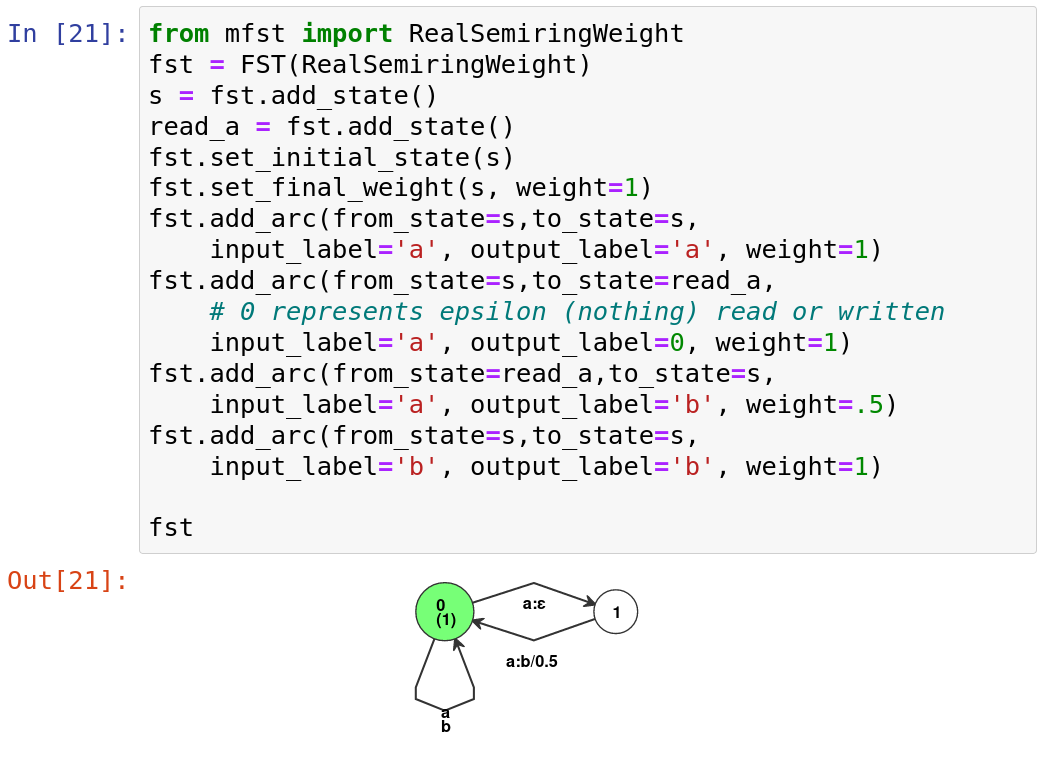}
  \caption{Here we define a weighted FST using the \texttt{RealSemiringWeight}
    $\langle +, * \rangle$.  This FST transduces between the alphabet
    $\{ \text{a}, \text{b} \}$ and itself.  Both symbols can be read unchanged
    (as they transverse the edge from state $0 \to 0$, which has weight 1).  If
    the FST encounters two \emph{``a''}s in a row, then it is transduced
    non-deterministically into the letter \emph{``b''} with a weight of $0.5$.  The
    weight $0.5$ comes from the fact that it multiplies the weights along the
    edges from state $0\to1$ and $1\to0$ when reading the sequence \emph{``aa.''}}
  \label{fig:double-a}
\end{figure}

\begin{figure}[H]
  \centering
  \includegraphics[width=.8\textwidth]{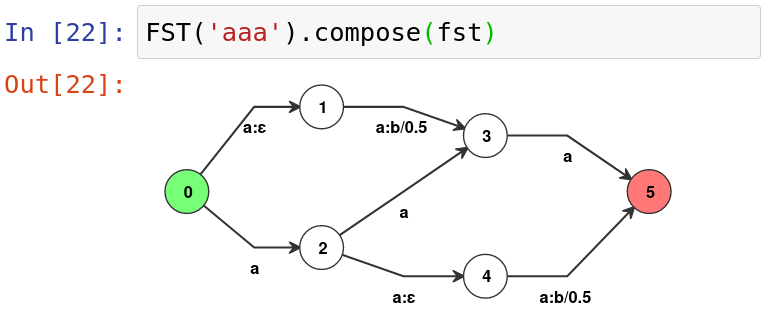}
  \caption{Here the FST defined in \cref{fig:double-a} is composed on its input
    side with the FSA which represents \emph{``aaa''} (constructed in the same way as
    \cref{fig:fsa-hello-code}).  The \texttt{FST('aaa')} is using the default
    semiring, so it is automatically cast to the RealSemiringWeight which was
    used to define \texttt{fst}.  This FST has three different paths which can
    generate outputs \emph{``aaa,''} \emph{``ba,''} or \emph{``ab.''}  The weight associated with
    each path is the result of multiplying the weights from both FSTs which are
    composed together.  The \texttt{FST('aaa')} only has a single path.
    The FST from \cref{fig:double-a} scores the \emph{``aa''}$\to$\emph{``b''}
    transduction with weight $0.5$.}
  \label{fig:compose-aaa}
\end{figure}

\subsection{Custom Weighted Semirings}

All semirings in \mfst are defined in pure Python by extending the
\href{https://github.com/matthewfl/openfst-wrapper/blob/dea1c93741b0c3be3dcedaa5d6f9158ce4126453/mfst/__init__.py#L16}{\texttt{AbstractSemiringWeight}}
class.  This means that we can easily define a custom semiring that makes use
of other libraries in Python.  For example, we may be interested in learning the
weights on a FST.  One way in which this can be done is to create a featurized
semiring, where we are tracking the weights associated with every feature
rather than the weight of an edge itself (\cref{sec:basic-featurized}).

\begin{figure}[h]
  \centering
  \includegraphics[width=\textwidth]{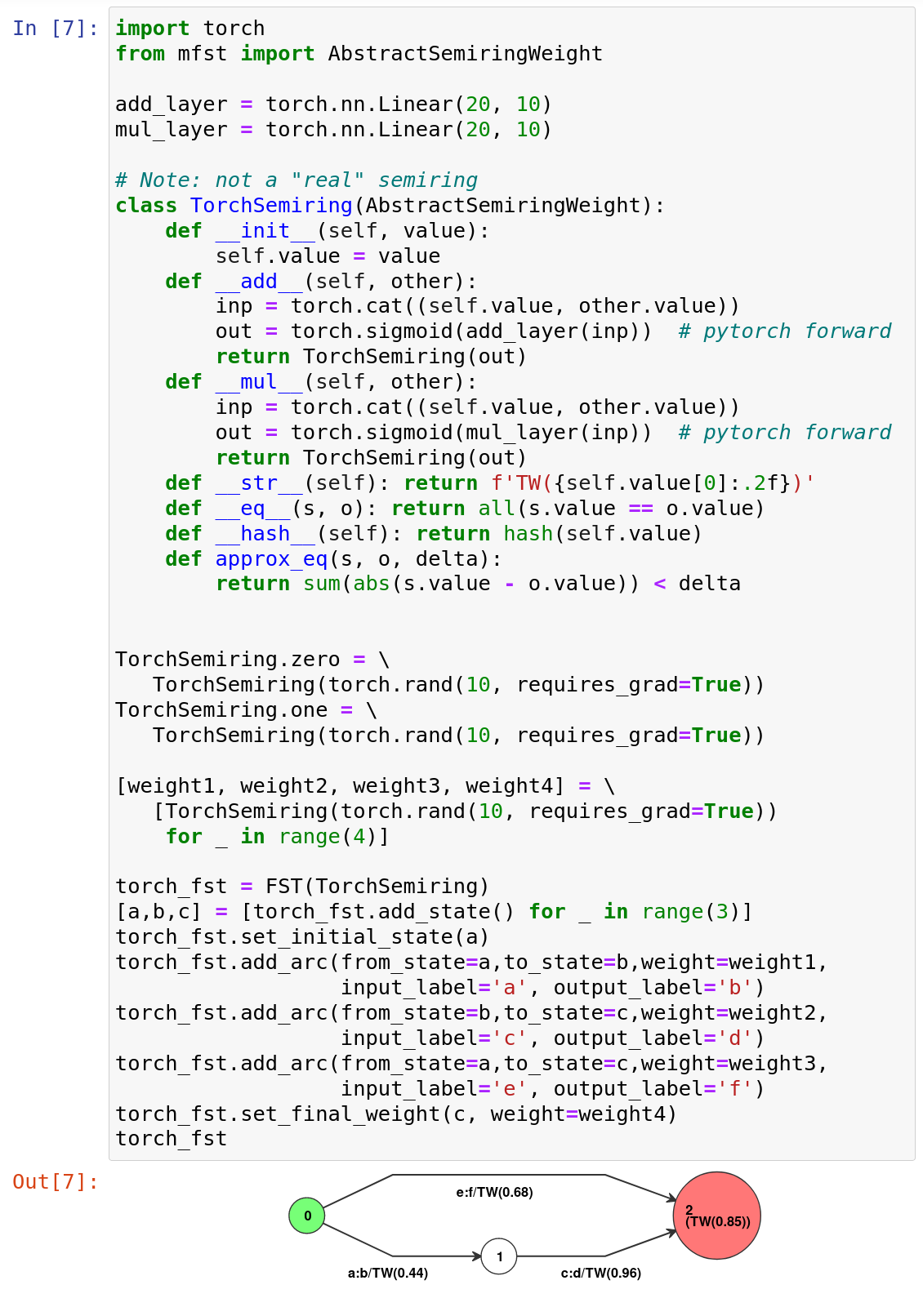}
  \caption{A custom semiring where weights are 10-element PyTorch tensors.  The
    multiplication and addition operations of the semiring are calling PyTorch Neural
    Network operators.  Note: This example is intended only to illustate
    that PyTorch tensors can be passed through OpenFST as weights.  The
    ``semiring'' defined here is not, in fact, a ``mathematically correct''
    semiring, and so actually using this as written here is ill-advised.}
  \label{fig:pytorch-semiring}
\end{figure}

Another approach that we can use for learning the weights on a FST is to
leverage a dynamic graph neural network frameworks, such as PyTorch. 
To construct a graph using standard neural network forward propagation, we simply use PyTorch's operations whenever we perform a semiring operation.
In \cref{fig:pytorch-semiring} we construct a custom
semiring that wraps a PyTorch tensors as a weight.  We can implement the
\texttt{\_\_add\_\_} and \texttt{\_\_mul\_\_} methods with anything which
results in a semiring.  For example, if we were to implement the min
semiring, then we would use \texttt{torch.min} for the \texttt{\_\_add\_\_}
method and use \texttt{torch.add} for the \texttt{\_\_mul\_\_} method.  For this example, however, we have instead chosen to exemplify how we could 
perform any neural operation.  Here we are using a forward pass through
a linear layer followed by a sigmoid operation.  Note, the
``semiring'' which we define in \cref{fig:pytorch-semiring} is not a \emph{real}
semiring as it does not follow the required distributivity and identity elements
of a semiring\footnote{\url{https://en.wikipedia.org/wiki/Semiring\#Definition}}.
While \mfst will still run, the outputs of an invalid ``semiring'' is unlikely
to be stable or work in practice.

\begin{figure}
  \centering
  \begin{subfigure}[t]{.45\textwidth}
    \includegraphics[width=\textwidth,clip,trim=0 0 2 0]{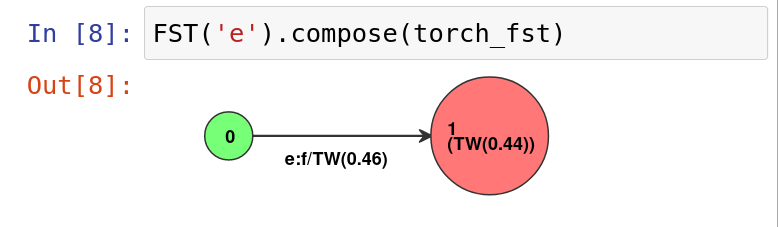}
    \caption{Composition with the default BooleanSemiring automatically casts to
      the identity elements of the TorchSemiring and then composes the two FSTs.}
  \end{subfigure}%
  \hfill
  \begin{subfigure}[t]{.45\textwidth}
    \vspace{-22mm}
    \includegraphics[width=\textwidth]{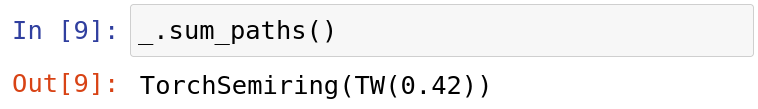}
    \vspace{7mm}
    \caption{\texttt{sum\_paths} sums all of the paths in the FST and returns
      the resulting weight element in the semiring.}
  \end{subfigure}
  \caption{\texttt{torch\_fst} manipulated through \mfst's built-in operations.}
\end{figure}

\begin{figure}
  \centering
  \includegraphics[width=.8\textwidth]{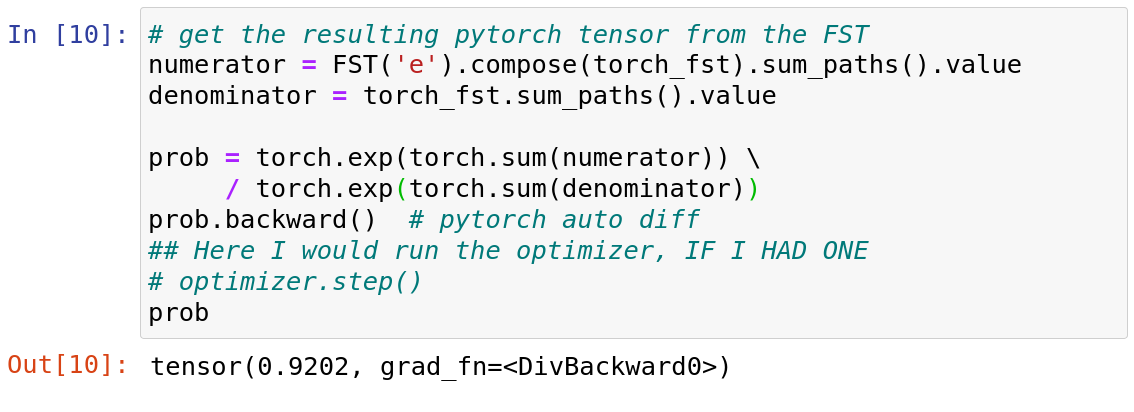}

  \caption{Here we exemplify how the \texttt{torch\_fst} could be used with
    PyTorch's backpropagation.  First, the numerator for the path that we are
    interested in is selected using \texttt{compose} and then
    \texttt{sum\_paths} to get the value for those specific paths.  We then get
    the normalizing constant (denominator) for all of the paths in the FST.  The
    variable \texttt{numerator} and \texttt{denominator} are both PyTorch
    tensors.  As a result, we can use PyTorch methods such as \texttt{torch.sum} and
    \texttt{torch.exp} on these variables.  As with a standard PyTorch tensor,
    we can use \texttt{tensor.backwards()} to compute the gradient.  In this
    example, we have included an optimizer, though that would be included in
    general.  Again, as mentioned above, this example is intended to illustrate how
    PyTorch tensors can be passed around through a FST as a semiring weight,
    not to illustrate something that is expected to work well.}
\end{figure}

\section{Implementation}

\mfst is a thin wrapper around OpenFST's C++ interface.  OpenFST~\cite{openfst}
has the ability to define custom semiring by defining custom C++
semiring
class\footnote{\url{http://www.openfst.org/twiki/bin/view/FST/FstAdvancedUsage\#Weights}}.
\mfst defines a custom semiring which wraps an opaque Python object.  The Python
C++ bridge makes use of PyBind11~\cite{pybind11}, which makes passing around
Python objects in C++ and exposing C++ methods to Python largely transparent.
Anytime an operation is performed by OpenFST on the semiring (such as addition
or multiplication), the call is redirected to the relevant method defined in Python.
This also means that we can easily define our own semiring from Python by
implementing the relevant methods.  This means that all FST operations
are handled by OpenFST, hence matured and well-tuned over the years.

The images that are drawn by \mfst make use of Jyputer's
\texttt{\_repr\_html\_}
method\footnote{\url{https://iPython.readthedocs.io/en/stable/config/integrating.html\#rich-display}}.
By defining a\linebreak
\texttt{\_repr\_html\_} method, this allows a class to define how it is rendered in a Jupyter
notebook by returning a string of HTML code that is used instead of the standard
textual output.  \mfst makes use of this method to generate HTML code that
utilizes D3.js~\cite{d3js} and Dagre-d3~\cite{dagre} to render the diagrams in
the browser.

\bibliography{bibfile}{}
\bibliographystyle{plain}

\clearpage
\appendix

\section{Abstract Semiring Class}

The AbstractSemiringWeight can be found in full detail here:
\url{https://git.io/JIWKn}.  Here is a shortened representation of the Abstract
Semiring base class.
\begin{lstlisting}[language=Python]
class AbstractSemiringWeight:
  semiring_properties : {'base', 'path'}          # if an idempotent(path) semiring
  def __add__(self, other): ...                   # semiring + operation (required)
  def __mul__(self, other): ...                   # semiring * operation (required)
  def __eq__(self, other): ...                    # if equal             (required)
  def __hash__(self): ...                         # hash code            (required)
  def __str__(self): ...                          # for displaying in figures
  def approx_eq(self, other, delta=1.0/1024): ... # approximately equal (suggested)

  def __div__(self, other): ...               # used by FST().push()
  def __pow__(self, n): ...
  def quantize(self, delta=1.0/1024): ...     # weight quantization
  def member(self): ...                       # if element of semiring (e.g. not nan)
  def reverse(self): ...                      # used by FST().reverse()
  def sampling_weight(self): ...              # used by FST().random_path()
  
AbstractSemiringWeight.zero = AbstractSemiringWeight(...)  # static 0 and 1 weights
AbstractSemiringWeight.one  = AbstractSemiringWeight(...)  # (required)
\end{lstlisting}

\section{Basic Featurized Semiring} \label{sec:basic-featurized}
This exhibits how a Python
\href{https://docs.python.org/3/library/collections.html#collections.Counter}{Counter}
could be used to track weights with a given feature.  Here the counts of a
feature are summed along a path, and we take the max value for a given feature in
the case that there are multiple accepting paths.

\begin{lstlisting}[language=Python]
from mfst import AbstractSemiringWeight
from collections import Counter
feature_weight = {}  # weights stored in a global dict
class FeaturizedWeight(AbstractSemiringWeight):
  def __init__(self, features):
    self._features = Counter(features)
    self._hash = hash(frozenset(self._features))
  def __add__(self, other):
    # take the max count for a given feature across different paths
    return FeaturizedWeight(self._features | other._features)
  def __mul__(self, other):
    return FeaturizedWeight(self._features + other._features)
  def __str__(self):
    return str(self._features)
  def __eq__(self, other):
    return self._hash == other._hash and self._features == other._features
  def __hash__(self):
    return self._hash
  def approx_eq(self, other, delta=1.0/1024):
    return sum(abs(self._features - other._features)) < delta
  def sampling_weight(self):
    result = 0
    for key, val in self._features.items():
      result += feature_weight.get(key, 0) * val
    return result
\end{lstlisting}

\end{document}